\documentclass[runningheads,a4paper]{llncs}

\usepackage{amsmath}
\usepackage{amssymb}
\usepackage[misc]{ifsym}
\usepackage{graphicx}
\usepackage{color}

\usepackage{multirow}
\usepackage{array}
\newcolumntype{L}[1]{>{\raggedright\let\newline\\\arraybackslash\hspace{0pt}}m{#1}}
\newcolumntype{C}[1]{>{\centering\let\newline\\\arraybackslash\hspace{0pt}}m{#1}}
\newcolumntype{R}[1]{>{\raggedleft\let\newline\\\arraybackslash\hspace{0pt}}m{#1}}

\usepackage{url}
\urldef{\mailsa}\path|{alfred.hofmann, ursula.barth, ingrid.haas, frank.holzwarth,|
\urldef{\mailsb}\path|anna.kramer, leonie.kunz, christine.reiss, nicole.sator,|
\urldef{\mailsc}\path|erika.siebert-cole, peter.strasser, lncs}@springer.com|

\begin{document}

\mainmatter  

\title{Deep Supervision for Pancreatic \\ Cyst Segmentation in Abdominal CT Scans}

\titlerunning{Deep Supervision for Pancreatic Cyst Segmentation in Abdominal CT Scans}

%
%
\author{Yuyin Zhou\textsuperscript{1}, Lingxi Xie\textsuperscript{2}$^{(\textrm{\Letter})}$,
Elliot K. Fishman\textsuperscript{3}, Alan L. Yuille\textsuperscript{4}}
\authorrunning{Y. Zhou {\em et al.}}

\institute{
\textsuperscript{1,2,4}The Johns Hopkins University, Baltimore, MD 21218, USA\\
\textsuperscript{3}The Johns Hopkins University School of Medicine, Baltimore, MD 21287, USA\\
\textsuperscript{1}{\tt\small zhouyuyiner@gmail.com}\quad
\textsuperscript{2}{\tt\small 198808xc@gmail.com}\\
\textsuperscript{3}{\tt\small efishman@jhmi.edu}\quad
\textsuperscript{4}{\tt\small alan.l.yuille@gmail.com}\\
\textcolor{red}{{\tt http://bigml.cs.tsinghua.edu.cn/\~{}lingxi/Projects/PanCystSeg.html}}
}

%
%

\toctitle{Deep Supervision for Pancreatic Cyst Segmentation in Abdominal CT Scans}
\tocauthor{Y. Zhou {\em et al.}}
\maketitle

\begin{abstract}
Automatic segmentation of an organ and its cystic region is a prerequisite of computer-aided diagnosis.
In this paper, we focus on pancreatic cyst segmentation in abdominal CT scan.
This task is important and very useful in clinical practice yet challenging due to the low contrast in boundary,
the variability in location, shape and the different stages of the pancreatic cancer.
Inspired by the high relevance between the location of a pancreas and its cystic region,
we introduce extra deep supervision into the segmentation network,
so that cyst segmentation can be improved with the help of relatively easier pancreas segmentation.
Under a reasonable transformation function, our approach can be factorized into two stages,
and each stage can be efficiently optimized via gradient back-propagation throughout the deep networks.
We collect a new dataset with $131$ pathological samples,
which, to the best of our knowledge, is the largest set for pancreatic cyst segmentation.
Without human assistance, our approach reports a $63.44\%$ average accuracy, measured by the Dice-S{\o}rensen coefficient (DSC),
which is higher than the number ($60.46\%$) without deep supervision.
\end{abstract}

\section{Introduction}
\label{Introduction}

In 2012, pancreatic cancers of all types were the 7th most common cause of cancer deaths,
resulting in $330\rm{,}000$ deaths globally~\cite{Stewart_2014_World}.
By the time of diagnosis, pancreatic cancer has often spread to other parts of the body.
Therefore, it is very important to use medical imaging analysis to assist identifying malignant cysts
in the early stages of pancreatic cancer to increase the survival chance of a patient~\cite{Dmitriev_2016_Pancreas}.
The emerge of deep learning has largely advanced the field of computer-aided diagnosis (CAD).
With the help of the state-of-the-art deep convolutional neural networks~\cite{Krizhevsky_2012_ImageNet}\cite{Simonyan_2015_Very},
such as the fully-convolutional networks (FCN)~\cite{Long_2015_Fully} for semantic segmentation,
researchers have achieved accurate segmentation on many abdominal organs.
There are often different frameworks for segmenting different organs~\cite{Al-Ayyoub_2013_Brain}\cite{Roth_2015_DeepOrgan}.
Meanwhile, it is of great interest
to find the lesion area in an organ~\cite{Wang_2016_Deep}\cite{Christ_2017_Automatic}\cite{Havaei_2017_Brain},
which, frequently, is even more challenging due to the tiny volume and variable properties of these parts.

This paper focuses on segmenting pancreatic cyst from abdominal CT scan.
Pancreas is one of the abdominal organs that are very difficult to be segmented
even in the healthy cases~\cite{Roth_2015_DeepOrgan}\cite{Roth_2016_Spatial}\cite{Zhou_2016_Pancreas},
mainly due to the low contrast in the boundary and the high variability in its geometric properties.
In the pathological cases, the difference in the pancreatic cancer stage
also impacts both the morphology of the pancreas and the cyst~\cite{Gintowt_2009_Unusual}\cite{Lasboo_2010_Morphological}.
Despite the importance of pancreatic cyst segmentation, this topic is less studied:
some of the existing methods are based on old-fashioned models~\cite{Klauss_2009_Automatic},
and a state-of-the-art approach~\cite{Dmitriev_2016_Pancreas} requires a bounding box of the cyst to be annotated beforehand,
as well as a lot of interactive operations throughout the segmentation process to annotate some voxels on or off the target.
These requirements are often unpractical when the user is not well knowledgable in medicine ({\em e.g.}, a common patient).
This paper presents the first system which produces reasonable pancreatic cyst segmentation
{\bf without human assistance} on the testing stage.

Intuitively, the pancreatic cyst is often closely related to the pancreas,
and thus segmenting the pancreas (relatively easier) may assist the localization and segmentation of the cyst.
To this end, we introduce deep supervision~\cite{Lee_2015_Deeply} into the original segmentation network,
leading to a joint objective function taking both the pancreas and the cyst into consideration.
Using a reasonable transformation function, the optimization process can be factorized into two stages,
in which we first find the pancreas, and then localize and segment the cyst based on the predicted pancreas mask.
Our approach works efficiently based on a recent published coarse-to-fine segmentation approach~\cite{Zhou_2016_Pancreas}.
We perform experiments on a newly collected dataset with $131$ pathological samples from CT scan.
Without human assistance on the testing stage,
our approach achieves an average Dice-S{\o}rensen coefficient (DSC) of $63.44\%$, which is practical for clinical applications.

\section{Approach}
\label{Approach}

\subsection{Formulation}
\label{Approach:Formulation}

Let a CT-scanned image be a 3D volume $\mathbf{X}$.
Each volume is annotated with ground-truth pancreas segmentation $\mathbf{P}^\star$ and cyst segmentation $\mathbf{C}^\star$,
and both of them are of the same dimensionality as $\mathbf{X}$.
${P_i^\star}={1}$ and ${C_i^\star}={1}$ indicate a foreground voxel of pancreas and cyst, respectively.
Denote a cyst segmentation model as $\mathbb{M}:{\mathbf{C}}={\mathbf{f}\!\left(\mathbf{X};\boldsymbol{\Theta}\right)}$,
where $\boldsymbol{\Theta}$ denotes the model parameters.
The loss function can be written as $\mathcal{L}\!\left(\mathbf{C},\mathbf{C}^\star\right)$.
In a regular deep neural network such as our baseline, the fully-convolutional network (FCN)~\cite{Long_2015_Fully},
we optimize $\mathcal{L}$ with respect to the network weights $\boldsymbol{\Theta}$ via gradient back-propagation.
To deal with small targets, we follow~\cite{Milletari_2016_VNet} to compute the DSC loss function:
${\mathcal{L}\!\left(\mathbf{C},\mathbf{C}^\star\right)}=
    {\frac{2\times{\sum_i}C_iC_i^\star}{{\sum_i}C_i+{\sum_i}C_i^\star}}$.
The gradient $\frac{\partial\mathcal{L}\!\left(\mathbf{C},\mathbf{C}^\star\right)}{\partial\mathbf{C}}$ can be easily computed.

\newcommand{\figurewidth}{9.0cm}
\begin{figure}[t]
\begin{center}
    \includegraphics[width=\figurewidth]{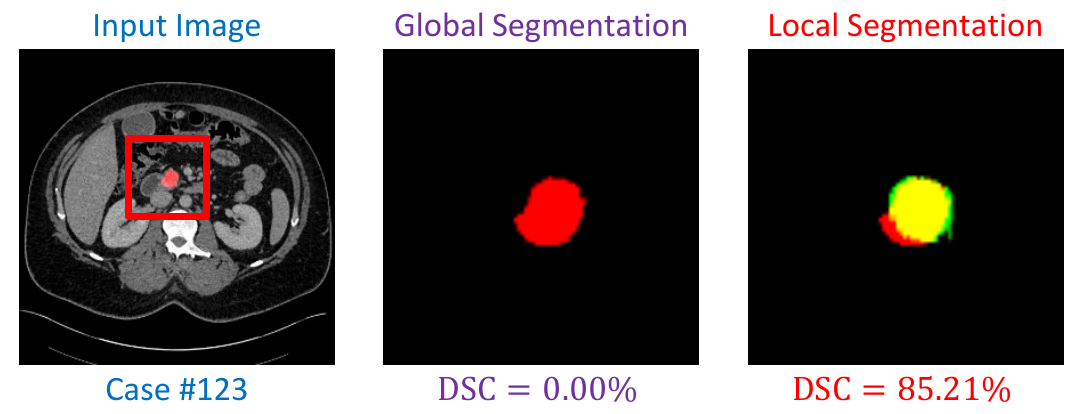}
\end{center}
\caption{
    A relatively difficult case in pancreatic cyst segmentation and the results produced by different input regions,
    namely using the entire image and the region around the ground-truth pancreas mask (best viewed in color).
    The cystic, predicted and overlapping regions are marked by red, green and yellow, respectively.
    For better visualization, the right two figures are zoomed in w.r.t. the red frame.
}
\label{Fig:Comparison}
\end{figure}

Pancreas is a small organ in human body, which typically occupies less than $1\%$ voxels in a CT volume.
In comparison, the pancreatic cyst is even smaller.
In our newly collected dataset, the fraction of the cyst, relative to the entire volume, is often much smaller than $0.1\%$.
In a very challenging case, the cyst only occupies $0.0015\%$ of the volume, or around $1.5\%$ of the pancreas.
This largely increases the difficulty of segmentation or even localization.
Figure~\ref{Fig:Comparison} shows a representative example
where cyst segmentation fails completely when we take the entire 2D slice as the input.

To deal with this problem, we note that the location of the pancreatic cyst is highly relevant to the pancreas.
Denote the set of voxels of the pancreas as ${\mathcal{P}^\star}={\left\{i\mid P_i^\star=1\right\}}$,
and similarly, the set of cyst voxels as ${\mathcal{C}^\star}={\left\{i\mid C_i^\star=1\right\}}$.
Frequently, a large fraction of $\mathcal{C}^\star$ falls within $\mathcal{P}^\star$
({\em e.g.}, ${\left|\mathcal{P}^\star\cap\mathcal{C}^\star\right|/\left|\mathcal{C}^\star\right|}>{95\%}$
in $121$ out of $131$ cases in our dataset).
Starting from the pancreas mask increases the chance of accurately segmenting the cyst.
Figure~\ref{Fig:Comparison} shows an example of using the ground-truth pancreas mask to recover the failure case of cyst segmentation.

This inspires us to perform cyst segmentation based on the pancreas region, which is relatively easy to detect.
To this end, we introduce the pancreas mask $\mathbf{P}$ as an explicit variable of our approach,
and append another term to the loss function to jointly optimize both pancreas and cyst segmentation networks.
Mathematically, let the pancreas segmentation model be
$\mathbb{M}_\mathrm{P}:{\mathbf{P}}={\mathbf{f}_\mathrm{P}\!\left(\mathbf{X};\boldsymbol{\Theta}_\mathrm{P}\right)}$,
and the corresponding loss term be $\mathcal{L}_\mathrm{P}\!\left(\mathbf{P},\mathbf{P}^\star\right)$.
Based on $\mathbf{P}$, we create a smaller input region by applying a transformation
${\mathbf{X}'}={\sigma\!\left[\mathbf{X},\mathbf{P}\right]}$, and feed $\mathbf{X}'$ to the next stage.
Thus, the cyst segmentation model can be written as
$\mathbb{M}_\mathrm{C}:{\mathbf{C}}={\mathbf{f}_\mathrm{C}\!\left(\mathbf{X}';\boldsymbol{\Theta}_\mathrm{C}\right)}$,
and we have the corresponding loss them $\mathcal{L}_\mathrm{C}\!\left(\mathbf{C},\mathbf{C}^\star\right)$.
To optimize both $\boldsymbol{\Theta}_\mathrm{P}$ and $\boldsymbol{\Theta}_\mathrm{C}$, we consider the following loss function:
\begin{equation}
\label{Eqn:LossFunction}
{\mathcal{L}\!\left(\mathbf{P},\mathbf{P}^\star,\mathbf{C},\mathbf{C}^\star\right)}=
    {\lambda\mathcal{L}_\mathrm{P}\!\left(\mathbf{P},\mathbf{P}^\star\right)+
    \left(1-\lambda\right)\mathcal{L}_\mathrm{C}\!\left(\mathbf{C},\mathbf{C}^\star\right)},
\end{equation}
where $\lambda$ is the balancing parameter defining the weight between either terms.

\subsection{Optimization}
\label{Approach:Optimization}

\renewcommand{\figurewidth}{11.5cm}
\begin{figure}[t]
\begin{center}
    \includegraphics[width=\figurewidth]{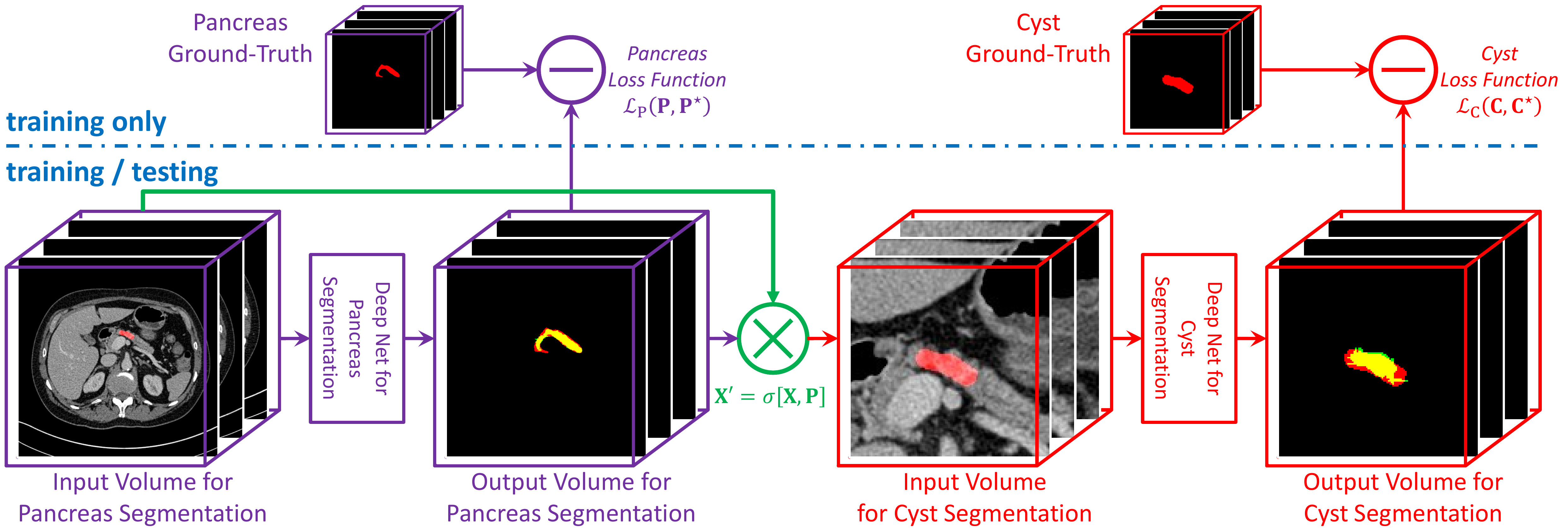}
\end{center}
\caption{
    The framework of our approach (best viewed in color).
    Two deep segmentation networks are stacked, and two loss functions are computed.
    The predicted pancreas mask is used in transforming the input image for cyst segmentation.
}
\label{Fig:Framework}
\end{figure}

We use gradient descent for optimization,
which involves computing the gradients over $\boldsymbol{\Theta}_\mathrm{P}$ and $\boldsymbol{\Theta}_\mathrm{C}$.
Among these, ${\frac{\partial\mathcal{L}}{\partial\boldsymbol{\Theta}_\mathrm{C}}}=
    {\frac{\partial\mathcal{L}_\mathrm{C}}{\partial\boldsymbol{\Theta}_\mathrm{C}}}$,
and thus we can compute it via standard back-propagation in a deep neural network.
On the other hand, $\boldsymbol{\Theta}_\mathrm{P}$ is involved in both loss terms, and applying the chain rule yields:
\begin{equation}
\label{Eqn:GradientP}
{\frac{\partial\mathcal{L}}{\partial\boldsymbol{\Theta}_\mathrm{P}}}=
    {\frac{\partial\mathcal{L}_\mathrm{P}}{\partial\boldsymbol{\Theta}_\mathrm{P}}+
    \frac{\partial\mathcal{L}_\mathrm{C}}{\partial\mathbf{X}'}\cdot
    \frac{\partial\mathbf{X}'}{\partial\mathbf{P}}\cdot
    \frac{\partial\mathbf{P}}{\partial\boldsymbol{\Theta}_\mathrm{P}}}.
\end{equation}
The second term on the right-hand side depends on the definition of ${\mathbf{X}'}={\sigma\!\left[\mathbf{X},\mathbf{P}\right]}$.
In practice, we define a simple transformation to simplify the computation.
The intensity value (directly related to the Hounsfield units in CT scan) of each voxel is either preserved or set as $0$,
and the criterion is whether there exists a nearby voxel which is likely to fall within the pancreas region:
\begin{equation}
\label{Eqn:Activation}
{X'_i}={X_i\times\mathbb{I}\!\left\{\exists j\mid{P_j>0.5}\wedge{\left|i-j\right|<t}\right\}},
\end{equation}
where $t$ is the threshold which is the farthest distance from a cyst voxel to the pancreas volume.
We set ${t}={15}$ in practice, and our approach is not sensitive to this parameter.
With this formulation, {\em i.e.}, ${\frac{\partial X_i'}{\partial P_j}}={0}$ almost everywhere.
Thus, we have ${\frac{\partial\mathbf{X}'}{\partial\mathbf{P}}}={\mathbf{0}}$ and
$\frac{\partial\mathcal{L}}{\partial\boldsymbol{\Theta}_\mathrm{P}}=
    {\frac{\partial\mathcal{L}_\mathrm{P}}{\partial\boldsymbol{\Theta}_\mathrm{P}}}$.
This allows us to factorize the optimization into two stages in both training and testing.
Since $\frac{\partial\mathcal{L}}{\partial\boldsymbol{\Theta}_\mathrm{P}}$ and
$\frac{\partial\mathcal{L}}{\partial\boldsymbol{\Theta}_\mathrm{C}}$ are individually optimized,
the balancing parameter $\lambda$ in Eqn~\eqref{Eqn:LossFunction} can be ignored.
The overall framework is illustrated in Figure~\ref{Fig:Framework}.
In training, we directly set ${\mathbf{X}'}={\sigma\!\left[\mathbf{X},\mathbf{P}^\star\right]}$,
so that the cyst segmentation model $\mathbb{M}_\mathrm{C}$ receives more reliable supervision.
In testing, starting from $\mathbf{X}$, we compute $\mathbf{P}$, $\mathbf{X}'$ and $\mathbf{C}$ orderly.
Dealing with two stages individually reduces the computational overheads.
It is also possible to formulate the second stage as multi-label segmentation.

The implementation details follow our recent work~\cite{Zhou_2016_Pancreas},
which achieves the state-of-the-art performance in the NIH pancreas segmentation dataset~\cite{Roth_2015_DeepOrgan}.
Due to the limited amount of training data, instead of applying 3D networks directly,
we cut each 3D volume into a series of 2D pieces, and feed them into a fully-convolutional network (FCN)~\cite{Long_2015_Fully}.
This operation is performed along three directions, namely the coronal, sagittal and axial views.
At the testing stage, the cues from three views are fused,
and each voxel is considered to fall on foreground if at least two views predict so.

In the pathological dataset, we observe a large variance in the morphology of both pancreas and cyst.
which increases the difficulty for the deep network to converge in the training process.
Consequently, using one single model may result in less stable segmentation results.
In practice, we use the FCN-8s model~\cite{Long_2015_Fully} from the pre-trained weights on the PascalVOC dataset.
This model is based on a $16$-layer VGGNet~\cite{Simonyan_2015_Very}, and we believe a deeper network may lead to better results.
We fine-tune it through $60\mathrm{K}$ iterations with a learning rate of $10^{-5}$.
Nine models, namely the snapshots after $\left\{20\mathrm{K},25\mathrm{K},\ldots,60\mathrm{K}\right\}$ iterations,
are used to test each volume, and the final result is obtained by computing the union of nine predicted foreground voxel sets.
Regarding other technical details, we simply follow~\cite{Zhou_2016_Pancreas},
including using the DSC loss layer instead of the voxel-wise loss layer
to prevent the bias towards background~\cite{Milletari_2016_VNet},
and applying a flexible convergence condition for fine-tuning at the testing stage.

\section{Experiments}
\label{Experiments}

\subsection{Dataset and Evaluation}
\label{Experiments:DatasetAndEvaluation}

We evaluate our approach on a dataset collected by the radiologists in our team.
This dataset contains $131$ contrast-enhanced abdominal CT volumes,
and each of them is manually labeled with both pancreas and pancreatic cyst masks.
The resolution of each CT scan is $512\times512\times L$,
where ${L}\in{\left[358,1121\right]}$ is the number of sampling slices along the long axis of the body.
The slice thickness varies from $0.5\mathrm{mm}$--$1.0\mathrm{mm}$.
We split the dataset into $4$ fixed folds, and each of them contains approximately the same number of samples.
We apply cross validation, {\em i.e.}, training our approach on $3$ out of $4$ folds and testing it on the remaining one.
We measure the segmentation accuracy by computing the Dice-S{\o}rensen Coefficient (DSC) for each 3D volume.
This is a similarity metric between the prediction voxel set $\mathcal{A}$ and the ground-truth set $\mathcal{G}$,
its mathematical form is ${\mathrm{DSC}\!\left(\mathcal{A},\mathcal{G}\right)}=
    {\frac{2\times\left|\mathcal{A}\cap\mathcal{G}\right|}{\left|\mathcal{A}\right|+\left|\mathcal{G}\right|}}$.
We report the average DSC score together with other statistics over all $131$ testing cases from $4$ testing folds.

\subsection{Results}
\label{Experiments:Results}

\newcommand{\colwidthA}{2.2cm}
\newcommand{\colwidthB}{2.2cm}
\begin{table}[t]
\centering
\begin{tabular}{|l||R{\colwidthA}|R{\colwidthA}|}
\hline
Method                                                   & Mean DSC                 & Max/Min DSC     \\
\hline\hline
Pancreas Segmentation, w/ GT Pancreas B-Box              & $83.99\pm 4.33$          & $93.82$/$69.54$ \\
\hline
{\bf Pancreas Segmentation}                              & $79.23\pm 9.72$          & $92.95$/$34.65$ \\
\hline\hline
Cyst Segmentation, w/ GT Cyst B-Box                      & $77.92\pm12.83$          & $96.14$/$24.69$ \\
\hline
{\bf Cyst Segmentation, w/o Deep Supervision}            & $60.46\pm31.37$          & $95.67$/$ 0.00$ \\
\hline
{\bf Cyst Segmentation, w/ Deep Supervision}             & $\mathbf{63.44}\pm27.71$ & $95.55$/$ 0.00$ \\
\hline
\end{tabular}
\caption{
    Pancreas and cyst segmentation accuracy, measured by DSC ($\%$), produced by different approaches.
    Bold fonts indicate the results that oracle information (ground-truth bounding box) is not used.
    With deep supervision, the average accuracy cyst segmentation is improved, and the standard deviation is decreased.
}
\label{Tab:Accuracy}
\end{table}

\subsubsection{Cystic Pancreas Segmentation.}
\label{Experiments:Results:Pancreas}

We first investigate pathological pancreas segmentation which serves as the first stage of our approach.
With the baseline approach described in~\cite{Zhou_2016_Pancreas}, we obtain an average DSC of $79.23\pm9.72\%$.
Please note that this number is lower than $82.37\pm5.68\%$,
which was reported by the same approach in the NIH pancreas segmentation dataset with $82$ healthy samples.
Meanwhile, we report $34.65\%$ DSC in the worst pathological case,
while this number is $62.43\%$ in the NIH dataset~\cite{Zhou_2016_Pancreas}.
Therefore, we can conclude that a cystic pancreas is more difficult to segment than a normal case.

\subsubsection{Pancreatic Cyst Segmentation.}
\label{Experiments:Results:Cyst}

Based on the predicted pancreas mask, we now study the pancreatic cyst segmentation which is the second stage in our approach.
Over $131$ testing cases, our approach reports an average DSC of $63.60\pm27.71\%$,
obtaining $2.98\%$ absolute or $4.92\%$ relative accuracy gain over the baseline.
The high standard deviation ($27.71\%$) indicates the significant variance in the difficulty of cyst segmentation.
On the one hand, our approach can report rather high accuracy ({\em e.g.}, $>95\%$ DSC) in some easy cases.
On the other hand, in some challenging cases, if the oracle cyst bounding box is unavailable,
both approaches (with or without deep supervision) can come into a complete failure ({\em i.e.}, DSC is $0\%$).
In comparison, our approach with deep supervision misses $8$ cyst cases, while the version without deep supervision misses $16$.

To the best of our knowledge, pancreatic cyst segmentation is very few studied previously.
A competitor is~\cite{Dmitriev_2016_Pancreas} published in 2016, which combines random walk and region growth for segmentation.
However, it requires the user to annotate the region-of-interest (ROI) beforehand,
and provide interactive annotations on foreground/background voxels throughout the segmentation process.
In comparison, when the bounding box is provided or not,
our approach achieves $77.92\%$ and $63.44\%$ average accuracies, respectively.
Being cheap or free in extra annotation, our approach can be widely applied to automatic diagnosis,
especially for the common users without professional knowledge in medicine.

\subsubsection{Visualization.}
\label{Experiments:Results:Visualization}

\renewcommand{\figurewidth}{11.0cm}
\begin{figure}[t]
\begin{center}
    \includegraphics[width=\figurewidth]{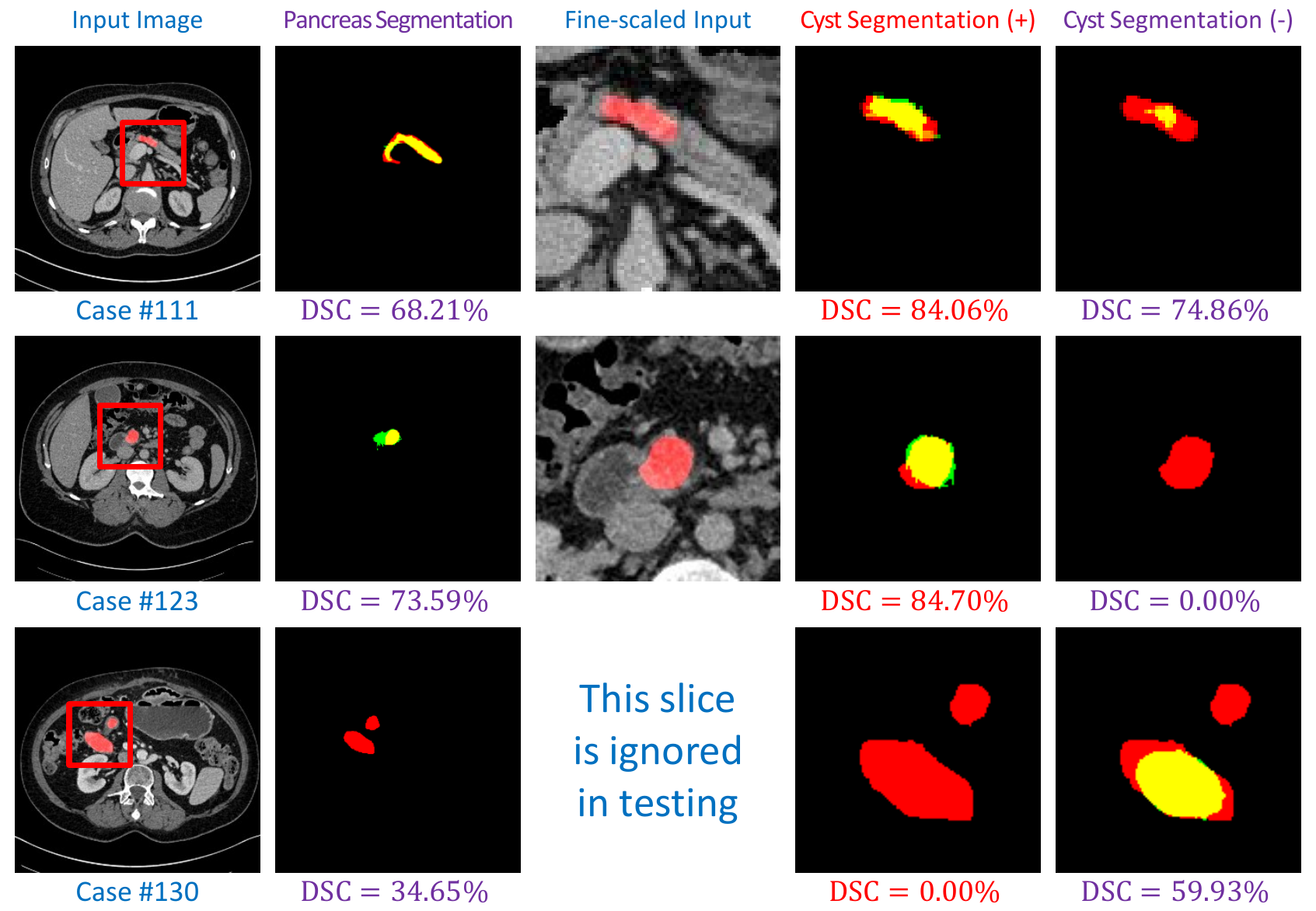}
\end{center}
\caption{
    Sample pancreas and pancreatic cyst segmentation results (best viewed in color).
    From left to right: input image (in which pancreas and cyst are marked in red and green, respectively),
    pancreas segmentation result, and cyst segmentation results when we apply deep supervision (denoted by +) or not (-).
    The figures in the right three columns are zoomed in w.r.t. the red frames.
    In the last example, pancreas segmentation fails in this slice, resulting in a complete failure in cyst segmentation.
}
\label{Fig:SegmentationResults}
\end{figure}

Three representative cases are shown in Figure~\ref{Fig:SegmentationResults}.
In the first case, both the pancreas and the cyst can be segmented accurately from the original CT scan.
In the second case, however, the cyst is small in volume and less discriminative in contrast,
and thus an accurate segmentation is only possible when we roughly localize the pancreas and shrink the input image size accordingly.
The accuracy gain of our approach mainly owes to the accuracy gain of this type of cases.
The third case shows a failure example of our approach,
in which an inaccurate pancreas segmentation leads to a complete missing in cyst detection.
Note that the baseline approach reports a $59.93\%$ DSC in this case,
and, if the oracle pancreas bounding box is provided, we can still achieve a DSC of $77.56\%$.
This inspires us that cyst segmentation can sometimes help pancreas segmentation, and this topic is left for future research.

\section{Conclusions}
\label{Conclusions}

This paper presents the first system for pancreatic cyst segmentation which can work without human assistance on the testing stage.
Motivated by the high relevance of a cystic pancreas and a pancreatic cyst,
we formulate pancreas segmentation as an explicit variable in the formulation,
and introduce deep supervision to assist the network training process.
The joint optimization can be factorized into two stages, making our approach very easy to implement.
We collect a dataset with $131$ pathological cases.
Based on a coarse-to-fine segmentation algorithm, our approach produces reasonable cyst segmentation results.
It is worth emphasizing that our approach does not require any extra human annotations on the testing stage,
which is especially practical in assisting common patients in cheap and periodic clinical applications.

This work teaches us that a lesion can be detected more effectively by considering its highly related organ(s).
This knowledge, being simple and straightforward, is useful in the future work in pathological organ or lesion segmentation.

\vspace{0.3cm}
\noindent
{\bf Acknowledgements.}
This work was supported by the Lustgarten Foundation for Pancreatic Cancer Research.
We thank Dr. Seyoun Park for enormous help.

\bibliographystyle{splncs03}
\bibliography{typeinst}
\end{document}